\documentclass[letterpaper]{article} 
\usepackage{aaai24}  
\usepackage{times}  
\usepackage{helvet}  
\usepackage{courier}  
\usepackage[hyphens]{url}  
\usepackage{graphicx} 
\usepackage{natbib}  
\usepackage{caption} 
\usepackage{algorithm}
\usepackage{algorithmic}
\usepackage{times}
\usepackage{latexsym}
\usepackage{amsmath}
\usepackage{amssymb}
\usepackage{subcaption}
\usepackage{svg}
\usepackage{multirow}

\usepackage{amsthm}
\newtheorem{proposition}{Proposition}
\usepackage{tabularx}
\newcommand{\Tau}{\mathbb{T}}
\usepackage{algorithm}
\usepackage{algorithmic}
\newtheorem{definition}{Definition}

\usepackage{newfloat}
\usepackage{listings}
\DeclareCaptionStyle{ruled}{labelfont=normalfont,labelsep=colon,strut=off} 
\floatstyle{ruled}
\newfloat{listing}{tb}{lst}{}
\floatname{listing}{Listing}
\title{HGE: Embedding Temporal Knowledge Graphs in a Product Space of Heterogeneous Geometric Subspaces}
\author {
    Jiaxin Pan,\textsuperscript{\rm 1}
    Mojtaba Nayyeri, \textsuperscript{\rm 1}
    Yinan Li \textsuperscript{\rm 1}
    Steffen Staab \textsuperscript{\rm 1,2}
}
\affiliations {
    \textsuperscript{\rm 1} University of Stuttgart, Stuttgart, Germany\\
    \textsuperscript{\rm 2} University of Southampton, Southampton, United Kingdom\\
    jiaxin.pan@ki.uni-stuttgart.de, mojtaba.nayyeri@ki.uni-stuttgart.de, \\yinan9721@gmail.com,
    steffen.staab@ki.uni-stuttgart.de
    
}

\usepackage{bibentry}

\begin{document}

\maketitle

\begin{abstract}
Temporal knowledge graphs represent temporal facts $(s,p,o,\tau)$ relating a subject $s$ and an object $o$ via a relation label $p$ at time $\tau$, where $\tau$ could be a time point or time interval. Temporal knowledge graphs may exhibit static temporal patterns at distinct points in time and dynamic temporal patterns between different timestamps. 
In order to learn a rich set of static and dynamic temporal patterns and apply them for inference, several embedding approaches have been suggested in the literature.
However, as most of them resort to single underlying embedding spaces, their capability to model all kinds of temporal patterns was severely limited by having to adhere to the geometric property of their one embedding space. 
We lift this limitation by an embedding approach that maps temporal facts into a product space of several heterogeneous geometric subspaces with distinct geometric properties, i.e.\  Complex,  Dual, and  Split-complex spaces. In addition, we propose a temporal-geometric attention mechanism to integrate information from different geometric subspaces conveniently according to the captured relational and temporal information. 
Experimental results on standard temporal benchmark datasets favorably evaluate our approach against state-of-the-art models.
\end{abstract}

\section{Introduction}
Knowledge Graphs (KGs)~\cite{hogan2021knowledge} model facts in real-world applications as directed edge-labeled graphs. Temporal KGs (TKGs) include timestamps to their facts in order to model the temporal validity of facts. Depending on the representational model, timestamps may represent time points or time intervals. For instance, a quadruple \textit{(Boris Johnson, IsPrimeministerOf, UK, [2019, 2022])} in a TKG represents the fact that Boris Johnson is the prime minister of UK between 2019 and 2022.  

\begin{figure}[t]
    \centering
    \includegraphics[width=0.48\textwidth]{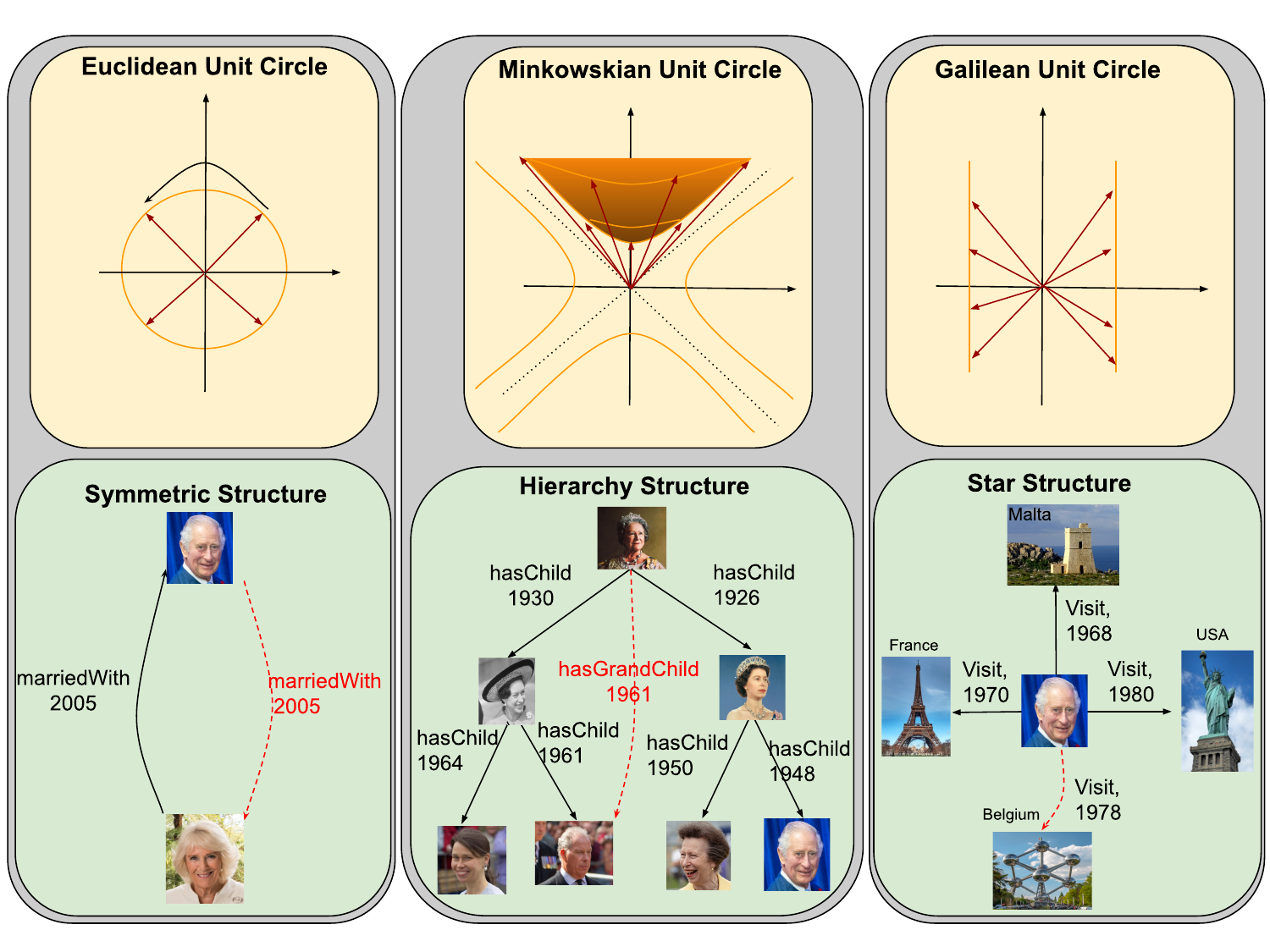}
    \caption{
    Unit spheres in their corresponding spaces. All points on the orange hyperplanes have the same distance to their origin. Different spaces favor different temporal patterns: 
    \textbf{Left:} Unit circle represented in Complex space (top) is suitable for representing periodicities and for inferencing with `periodic' logical temporal patterns, e.g.\ symmetry (bottom). 
    \textbf{Middle:} Minkowskian unit circle  in Split-complex space (top) is suitable for representing a temporal hierarchy formed by \textit{Make statement}.
    \textbf{Right:} Galilean unit circle represented in Dual space (top) is suitable for representing temporal star patterns (bottom).
    }
    \label{fig:baseMotivExamp}
\end{figure}

Relations in temporal knowledge graphs may exhibit various structural temporal patterns. In the left part of Figure \ref{fig:baseMotivExamp}, \textit{(Charles III, marriedWith, Camilla, 2005)} and \textit{(Camilla, marriedWith, Charles III, 2005)} forms a symmetrical structure in time.
In the middle part, at first \textit{(Elizabeth Bowes-Lyon, hasChild, Elizabeth II, 1926)} and then \textit{(Elizabeth II, hasChild, Charles III, 1948)}. The transition of \textit{hasChild} relation through \textit{Elizabeth II} forms a hierarchy structure in TKGs.
In the right part, \textit{Charles III}, \textit{Visit} \textit{Malta}, \textit{France, Belgium}, \textit{USA} etc at different timestamps, forming a star structure over time.
Moreover, as \textit{Charles III} shows, the structures which entities are involved in temporal knowledge graphs may evolve over time. 
How to preserve different relational structural patterns and how to capture evolving temporal patterns for entities is a fundamental challenge in TKGEs. 

Existing embedding approaches such as TeRO, RotateQVS, and TLT-KGE\cite{xu2020tero,chen2022rotateqvs,tltcomplexzhang2022along} resorted to single underlying embedding spaces, such as Complex space or Quaternion space to model symmetric patterns by the rotations on a unit hypersphere. Other works \cite{chami2020low,balazevic2019multi,herculemontella2021hyperbolic,han2020dyernie} use hyperbolic space to preserve hierarchical patterns in temporal KGs. However, their capability to model all kinds of structural patterns was severely limited by having to adhere to the geometric properties of their one embedding space. 
\cite{han2020dyernie} has shown the advantage of using multiple geometric subspaces (spherical, hyperbolic, etc) in different dimensions to preserve heterogeneous structural patterns in temporal KGs. However, it ignores the evolution of structural patterns between entities and requires a manual selection of subspaces dimension. How to integrate suitable subsets of geometries to model different relational structural patterns as well as capturing evolutionary temporal patterns between entities remain an open problem in these approaches. 
 

In this paper, we address these problems by introducing a new product space covering various geometric subspaces namely a) complex, b) split-complex and c) dual spaces with a \emph{temporal relational attention mechanism} and a \emph{temporal geometric attention mechanism} to model both structural and evolutionary temporal patterns. Figure \ref{fig:baseMotivExamp} illustrates the spaces and some corresponding patterns.
a) Consider the left part of Figure \ref{fig:baseMotivExamp}: 
In the complex space, Euclidean unit circles are induced by circular rotations.
Thus, points on the circle establish periodicities and various logical temporal patterns, e.g.\ relations that are symmetry in time \cite{xu2020tero}. 
Circular rotations are modeled by circular sine and cosine functions in the complex space. 
b) Consider the middle part of Figure \ref{fig:baseMotivExamp}: 
In the split-complex space, a Minkowskian unit circle is induced through hyperbolic rotation, where points on the circle can be mapped using hyperbolic sine and cosine.
Thus, the split-complex space can capture a temporal hierarchy, e.g.\ children must be born after their parents.
c) Consider the right part of Figure \ref{fig:baseMotivExamp}:
In the dual space, a Galilean unit circle is induced by the rotation that maps points on the circle using Galilean sine and cosine.
Points on the induced circle (two parallel lines) are equidistant to the center, making it useful for modeling star-shaped subgraphs.

The combination of these three spaces together with their geometries and corresponding operators allows for capturing diverse logical and structural patterns such as relational symmetry in time, temporal hierarchy patterns, and temporal star patterns. 
Which geometry should be preferred in a specific case, however, needs to be learned.
For this purpose, we provide a temporal geometric attention mechanism to select the preferred geometries for a given relation and time. 
Moreover, to deal with the evolution of patterns between entities, we propose the temporal-relational attention mechanism to balance static embedding and time-evolving embedding. 
We compare our TKGE model, heterogeneous geometric embedding (HGE), to TKGE methods in Complex space such as TComplEx \cite{tcomplexlacroix2020tensor}, TeRo \cite{xu2020tero}, TLT-KGE \cite{tltcomplexzhang2022along} and find that our model obtains better results for link prediction tasks in TKGs.
In summary, the key contributions of this paper are as follows:
\begin{itemize}
    \item We extend state-of-the-art Temporal Knowledge Graph Embedding (TKGE) models that use Complex spaces to a new method, HGE. By utilizing multiple heterogeneous geometries, HGE embeds temporal facts in a product space of Complex, Split-complex, and Dual subspaces. 
    \item Our theoretical analysis shows that our embedding method can capture a range of various structural and logical temporal patterns by utilizing the rotation operations acting on  Euclidean, Minkowskian, and Galilean unit circles.
    These theoretical considerations are supported by experiments and ablation studies on pre-existing benchmark datasets.
    \item Two novel kinds of attention mechanisms, temporal-relational attention, and temporal-geometric attention allow for representing relation changing frequencies and suitable geometries, respectively.
    \item Experimental results on benchmark datasets show that HGE uniformly improves several state-of-the-art TKGE models. Subsequent ablation studies verify the general benefit of the attention-based product space models over the Complex space.
\end{itemize} 

\section{Preliminaries}

\begin{definition}[Time Interval]
Let $\Tau$ be the set of closed intervals on the real line $\mathbb{R}$. For a time interval  $\tau=[m,n] \in\Tau, \tau\subseteq \mathbb{R}$, with $m,n\in \tau, m\leq n$ it holds that $\forall t\in \mathbb{R}: m\leq t\leq n\Rightarrow t\in\tau $.
\end{definition}


\begin{definition}[Temporal Knowledge Graph]
Let $V$ be a set of vertices, $R$ be a set of relation labels, $\Tau$ be the set of all time intervals,
$G\subseteq V\times R \times V \times \Tau$, then a temporal fact $(s,p,o,\tau)\in G$ with subject $s$, object $o$ and relation label $p$ is valid during time interval $\tau$. A temporal knowledge graph $TKG=(V,R,G)$ defines a set of temporal facts. In addition, 
we denote $G_i$ as $i$-th snapshot of the TKG
\end{definition}

We re-use Allen’s interval calculus to express relations between time intervals \cite{allen1983maintaining}. It defines 13 possible relations between two time intervals such that these relations are exhaustive and pairwise disjoint. For example, Allen relation Contains($\tau_1, \tau_2$) holds between two time intervals $\tau_1=[m_1,n_1], \tau_2=[m_2,n_2]$ if $m_1<m_2<n_2<n_1$. Following \cite{singh2023neustip}, we refer to the 13 relations of Allen interval calculus as \textit{Allen relations} and the relation in temporal knowledge graphs as \textit{KG relations}. Appendix A describes the details of 13 Allen relations.

\section{Embedding Model in Heterogeneous Geometric Subspaces}
\begin{figure*}[t]
    \centering
    \includegraphics[width=\textwidth]{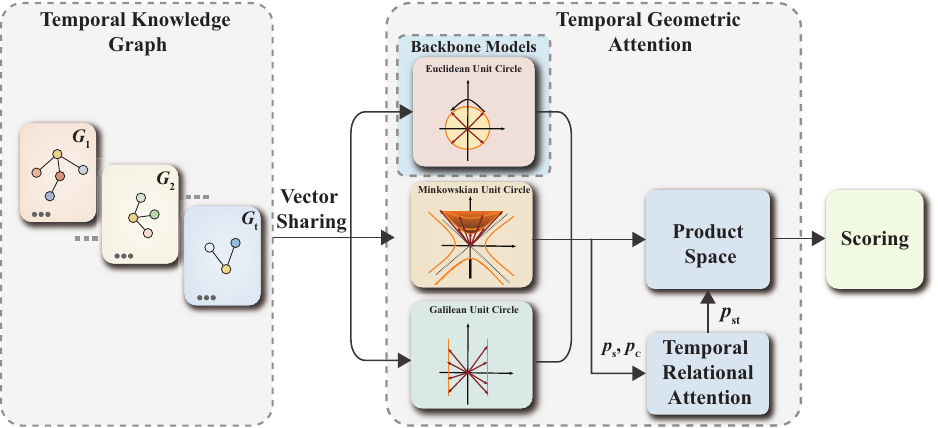}
    \caption{
An illustration for the HGE. At first, entities, relations and timestamps in temporal knowledge graphs are represented in heterogeneous geometric subspaces: 1) complex space, 2) split-complex space, 3) dual space respectively. Based on the static relation embedding $p_{s}$, and dynamic relation embedding $p_{c}$, temporal relational attention learns hybrid relation embedding $p_{st}$ based on each relation's changing frequencies. Temporal geometric attention incorporates embeddings in geometric subspaces into a product space by $p_{st}$, which decides the suitable geometry for each relation. Finally, the scoring function is performed on the embeddings learned in the product space.  
    }
    \label{fig:HGE_model}
\end{figure*}

To capture heterogeneous structural and logical patterns in a temporal KG, we propose the HGE model which extends the complex space adopted by existing models\cite{tltcomplexzhang2022along, tcomplexlacroix2020tensor} to an attention-based product space. We introduce the key components of our temporal knowledge graph embedding method, HGE, in the following order: 
a) \emph{embedding space},
b) \emph{temporal-relational attention}, 
c) \emph{temporal-geometric attention}. Figure \ref{fig:HGE_model} shows the structure of our proposed HGE model.

\subsection{\textbf{Embeddings in Geometric Subspaces}}
We aim to embed the elements of a temporal knowledge graph (entities, relations, and times) into a $d$ dimensional product space $\mathcal{M} = \mathbb{M}_1 \times \ldots \times \mathbb{M}_d$ where each $\mathbb{M}_i$ is a Complex, Dual or Split-complex space,
i.e.\ $\mathbb{M}_i \in \{ \mathbb{C}, \mathbb{S}, \mathbb{D} \}$.
For a given fact $(s,p,o,\tau)\in G$, we use the mappings $f_e: \mathcal{E} \xrightarrow{} \mathbb{M}_i, f_r: \mathcal{R} \xrightarrow{} \mathbb{M}_i, f_{\tau}: \mathcal{T} \xrightarrow{} \mathbb{M}_i$ to assign $d$ dimensional vectors to each element of a TKG as $\boldsymbol{s_{\mathbb{M}_i},p_{\mathbb{M}_i},o_{\mathbb{M}_i}, \tau_{\mathbb{M}_i}}$ respectively. 

We introduce the three fundamental parts of the product space for developing our model, namely Complex, Split-complex and Dual spaces together with their geometric interpretations.
Given a quadratic formula $k^2 + g = 0, g = \{-1, 1, 0\}$, we have the three number systems based on the value of $g$:

\paragraph{\textbf{Complex Vector Space}} 
Complex numbers \cite{harkin2004geometry,metahelzer2000special} allow for solving the quadratic formula $k^2 + 1 = 0,$ by defining a new number $k = i$ where $i^2 = -1$. 
$i$ is used to define the set of Complex numbers $\mathbb{C} = \{q = a + b i | a, b \in \mathbb{R}, i^2 = -1\}$, where $a$ is the real and $b$ the imaginary part. The multiplication of two Complex numbers $q_1= a+bi, q_2=c+di$ is defined by $q_1 * q_2 = (ac- bd) + (ad + bc) i$. It has been proved by previous works \cite{tltcomplexzhang2022along,tcomplexlacroix2020tensor,xu2020tero} to represent temporal knowledge graphs effectively. Following their work, we represent $\boldsymbol{s,p,o, \tau}$ in complex space as:

\begin{equation}
\label{eq:complex}
\begin{aligned}
        \boldsymbol{s_{\mathbb{C}}}=\boldsymbol{s}_{\mathbb{C}a} + \boldsymbol{s}_{\mathbb{C}b}i, \boldsymbol{p_{\mathbb{C}}}=\boldsymbol{p}_{\mathbb{C}a} + \boldsymbol{p}_{\mathbb{C}b}i,  \\ \boldsymbol{o_{\mathbb{C}}}=\boldsymbol{o}_{\mathbb{C}a} + \boldsymbol{o}_{\mathbb{C}b}i, \boldsymbol{\tau_{\mathbb{C}}}=\boldsymbol{\tau}_{\mathbb{C}a} + \boldsymbol{\tau}_{\mathbb{C}b}i
\end{aligned}
\end{equation}

where $\boldsymbol{s}_{\{.\}}, \boldsymbol{p}_{\{.\}}, \boldsymbol{o}_{\{.\}}, \boldsymbol{\tau}_{\{.\}} \in \mathbb{R}^{d}$. ${\{.\}}_{a}$ represents the real part of each element and ${\{.\}}_{b}$ represents the imaginary part.
 
\paragraph{\textbf{Split-complex Vector Space}} 
Dealing with quadratic formula $k^2 - 1 = 0,$ a split-complex number \cite{harkin2004geometry,metahelzer2000special} is defined as $p = a + j b$, where $k = j, j^2 = 1, j \neq 1, -1.$ Formally the space of split-complex number is defined as $\mathbb{S} = \{ q = a + bj | a,b \in \mathbb{R}, j^2 = 1, j \neq 1, -1\}$. $a,b$ are real and split parts, respectively. The multiplication of two Split-Complex numbers $q_1= a+bj, q_2=c+dj$ is defined by $q_1 * q_2 = (ac+ bd) + (ad + bc) j$. We represent $\boldsymbol{s,p,o, \tau}$ in split-complex space as:

\begin{equation}
\label{eq:split}
\begin{aligned}
        \boldsymbol{s_{\mathbb{S}}}=\boldsymbol{s}_{\mathbb{S}a} + \boldsymbol{s}_{\mathbb{S}b}j, \boldsymbol{p_{\mathbb{S}}}=\boldsymbol{p}_{\mathbb{S}a} + \boldsymbol{p}_{\mathbb{S}b}j,  \\ 
        \boldsymbol{o_{\mathbb{S}}}=\boldsymbol{o}_{\mathbb{S}a} + \boldsymbol{o}_{\mathbb{S}b}j,
        \boldsymbol{\tau_{\mathbb{S}}}=\boldsymbol{\tau}_{\mathbb{S}a} + \boldsymbol{\tau}_{\mathbb{S}b}j,
\end{aligned}
\end{equation}

\paragraph{\textbf{Dual Vector Space}} 
Dual numbers \cite{dualangeles1998application,metahelzer2000special} are  similar to Complex numbers, 
but their imaginary $\epsilon$ is defined such that $\epsilon^2 = 0, \epsilon \neq 0$.
The dual space is then defined as $\mathbb{D} = \{q = a + b \epsilon | a,b \in \mathbb{R}, \epsilon^2 = 0, \epsilon \neq 0\}$ where $a,b$ are real and dual components of the dual numbers. The multiplication of two Dual numbers $q_1= a+b\epsilon, q_2=c+d\epsilon$ is defined by $q_1 * q_2 = (ac) + (ad + bc) \epsilon$. We represent $\boldsymbol{s,p,o, \tau}$ in dual space as:

\begin{equation}
\label{eq:dual}
\begin{aligned}
        \boldsymbol{s_{\mathbb{D}}}=\boldsymbol{s}_{\mathbb{D}a} + \boldsymbol{s}_{\mathbb{D}b}\epsilon, \boldsymbol{p_{\mathbb{D}}}=\boldsymbol{p}_{\mathbb{D}a} + \boldsymbol{p}_{\mathbb{D}b}\epsilon,  \\ \boldsymbol{o_{\mathbb{D}}}=\boldsymbol{o}_{\mathbb{D}a} + \boldsymbol{o}_{\mathbb{D}b}\epsilon, \boldsymbol{\tau_{\mathbb{D}}}=\boldsymbol{\tau}_{\mathbb{D}a} + \boldsymbol{\tau}_{\mathbb{D}b}\epsilon
\end{aligned}
\end{equation}

\subsection{\textbf{Embeddings in Attention-based Product Space}}
How to fuse information from different subspaces into a product space  efficiently remains a challenging task in the knowledge graph embedding task. Existing work \cite{han2020dyernie} assigns different dimensions $d_{i}$ for each subspace $\mathbb{M}_i$, where $\sum{d_{i}} = d $, and calculates their individual loss which will be aggregated subsequently to a total loss. Such a stacking strategy requires the manual selection of suitable $d_{i}$ numbers for every new task and consumes huge computation resources to reach optimal $d_{i}$ decision. 
To capture suitable geometries from various subspaces efficiently, we introduce an attention-based product space. Rather than stacking ad hoc vectors for each subspace, our method reuses vectors for every subspace and aggregates \textit{Scoring Vectors} of subspaces by relational and temporal information.  

\paragraph{\textbf{Real and Imaginary Vector Sharing}} Existing methods \cite{han2020dyernie} assigns different vectors for each subspaces. However, pre-experiments in Appendix C illustrate that although their geometric interpretations are diverse, real and imaginary vectors in different subspaces are almost unanimous when trained to optimal settings with the same embedding sizes. Accordingly, we share the real and imaginary vectors between all subspaces as follows:

\begin{equation}
    \begin{split}
        &\{.\}_{\mathbb{C}a}=\{.\}_{\mathbb{S}a}=\{.\}_{\mathbb{D}a}\\
        &\{.\}_{\mathbb{C}b}=\{.\}_{\mathbb{S}b}=\{.\}_{\mathbb{D}b}\\
    \end{split}
\end{equation}
where $\{.\} \in \{s, p, o, \tau \}$. With the reusing strategy, our method avoids the manual selection of subspace dimensions and saves embedding space. If not specified, we use $\boldsymbol{s}=[\boldsymbol{s}_{a},\boldsymbol{s}_{b}]$, $\boldsymbol{p}=[\boldsymbol{p}_{a}, \boldsymbol{p}_{b}]$, $\boldsymbol{o}=[\boldsymbol{o}_{a},\boldsymbol{o}_{b}]$ and $\boldsymbol{\tau}=[\boldsymbol{\tau}_{a},\boldsymbol{\tau}_{b}]$ to represent embeddings a generic geometric subspace in the following section for simplicity.

\paragraph{\textbf{Temporal-relational Attention}}
Relations in TKGs may exhibit different
frequencies of change varying from fully static to quickly changing behavior \cite{tcomplexlacroix2020tensor}. 
For example, the relation \textit{capitalOf} is not changing often over time, while the relation \textit{isPresidentOf} exhibits more frequent changes. 
Therefore, for each relation $\boldsymbol{p}$, we provide two vectors $\boldsymbol{p}_s, \boldsymbol{p}_{c} \in \mathcal{M}$. The first captures the static behavior and the second captures the dynamic behavior by multiplication with time embedding $\boldsymbol{\tau}_\tau$. We provide a temporal attention mechanism to emphasize static or dynamic behavior depending on the characteristics of the relation:

\begin{equation}
\begin{split}
    \boldsymbol{p}_{s\tau} &= \alpha_\tau \left(\boldsymbol{p_c} * \boldsymbol{\tau}_\tau\right) + \alpha_s \boldsymbol{p}_s \\ 
    \left(\alpha_\tau, \alpha_s \right)&=  \operatorname{Softmax}\left(\mathbf{w}_p \left(\boldsymbol{p_c} * \boldsymbol{\tau}_\tau \right) , \mathbf{w}_p \boldsymbol{p}_s \right)
    \label{eq:temporalAttention}
\end{split}
\end{equation}

\noindent
where $\mathbf{w}_p$ is the relation-specific weight.

\paragraph{\textbf{Scoring Vectors from Subspaces}} 
We take all values in each subspace for entities, relations, and times
$\boldsymbol{s}_i, \boldsymbol{p}_{s\tau i}, \boldsymbol{o}_i \in \mathbb{M}_i$ and compute $ \boldsymbol{c}_i =  \langle \boldsymbol{s}_i, \boldsymbol{p}_{s\tau i}, {\boldsymbol{o}_i} \rangle$\footnote{Similar to previous work\cite{xu2020tero,lacroix2019tensor}, we adopt conjugate on $\boldsymbol{o}_i$ to increase the performance in experiments.}, where
$\langle , , \rangle$ is the product in Complex, Split-complex and Dual spaces computed as follows:
\begin{equation}
\begin{split}
   \boldsymbol{c}_\mathbb{C} & =  \langle (\boldsymbol{s}_{a} \boldsymbol{p}_{s \tau a} - \boldsymbol{s}_{b} \boldsymbol{p}_{s \tau b}) + (\boldsymbol{s}_{a} \boldsymbol{p}_{s \tau b} + \boldsymbol{s}_{b} \boldsymbol{p}_{s \tau a}) i, \boldsymbol{o}_{a} + i \boldsymbol{o}_{b} \rangle \\
    &= (\boldsymbol{s}_{a} \boldsymbol{p}_{s \tau a} \boldsymbol{o}_{a} - \boldsymbol{s}_{b} \boldsymbol{p}_{s \tau b} \boldsymbol{o}_{a} - \boldsymbol{s}_{a} \boldsymbol{p}_{s \tau b} \boldsymbol{o}_{b} - \boldsymbol{s}_{b} \boldsymbol{p}_{s \tau a} \boldsymbol{o}_{b}) + \\ 
    & (\boldsymbol{s}_{a} \boldsymbol{p}_{s \tau a} \boldsymbol{o}_{b} - \boldsymbol{s}_{b} \boldsymbol{p}_{s \tau b} \boldsymbol{o}_{b} + \boldsymbol{s}_{a} \boldsymbol{p}_{s \tau b} \boldsymbol{o}_{a} + \boldsymbol{s}_{b} \boldsymbol{p}_{s \tau a} \boldsymbol{o}_{a}) i, \\
    \boldsymbol{c}_\mathbb{S} &  = 
    \langle (\boldsymbol{s}_{a} \boldsymbol{p}_{s \tau a} + \boldsymbol{s}_{b} \boldsymbol{p}_{s \tau b}) + (\boldsymbol{s}_{a} \boldsymbol{p}_{s \tau b} + \boldsymbol{s}_{b} \boldsymbol{p}_{s \tau a}) j, \boldsymbol{o}_{a} + j \boldsymbol{o}_{b} \rangle \\
    &= (\boldsymbol{s}_{a} \boldsymbol{p}_{s \tau a} \boldsymbol{o}_{a} + \boldsymbol{s}_{b} \boldsymbol{p}_{s \tau b} \boldsymbol{o}_{a} + \boldsymbol{s}_{a} \boldsymbol{p}_{s \tau b} \boldsymbol{o}_{b} + \boldsymbol{s}_{b} \boldsymbol{p}_{s \tau a} \boldsymbol{o}_{b}) 
    + \\
    &(\boldsymbol{s}_{a} \boldsymbol{p}_{s \tau a} \boldsymbol{o}_{b} + \boldsymbol{s}_{b} \boldsymbol{p}_{s \tau b} \boldsymbol{o}_{b} + \boldsymbol{s}_{a} \boldsymbol{p}_{s \tau b} \boldsymbol{o}_{a} + \boldsymbol{s}_{b} \boldsymbol{p}_{s \tau a} \boldsymbol{o}_{a}) j,\\
         \boldsymbol{c}_\mathbb{D}  & = 
    \langle (\boldsymbol{s}_{a} \boldsymbol{p}_{s \tau a}) + (\boldsymbol{s}_{a} \boldsymbol{p}_{s \tau b} + \boldsymbol{s}_{b} \boldsymbol{p}_{s \tau a}) \epsilon, \boldsymbol{o}_{a} + \epsilon \boldsymbol{o}_{b} \rangle \\
    &= (\boldsymbol{s}_{a} \boldsymbol{p}_{s \tau a} \boldsymbol{o}_{a}) + (\boldsymbol{s}_{a} \boldsymbol{p}_{s \tau a} \boldsymbol{o}_{b} + \boldsymbol{s}_{a} \boldsymbol{p}_{s \tau b} \boldsymbol{o}_{a} + \boldsymbol{s}_{b} \boldsymbol{p}_{s \tau a} \boldsymbol{o}_{a}) j.
\end{split}
\end{equation}
\paragraph{Temporal-geometric Attention}
Scoring vectors represent distinctive geometric information captured by each subspace. We propose a temporal-geometric attention mechanism to integrate them based on current relational and time information.

\begin{equation}
\beta_{i}=\operatorname{Softmax}\left(\boldsymbol{p}_{s\tau} \boldsymbol{c}_i \right), \,\, i \in \{\mathbb{C}, \mathbb{D}, \mathbb{S}\}.
\label{eq:temporalAttention3}
\end{equation}


It emphasizes the most suitable geometry for each query via the augmented relation embedding $\boldsymbol{p}_{st}$. As the changing frequencies of relations could be reflected by $\boldsymbol{p}_{st}$, HGE could model the static and dynamic logical and structural patterns in TKGs.
The overall score aggregates the inner product in all subspaces:



\begin{equation}
S_{\mathcal{M}}(s,r,o, \tau) = \sum_{i=1}^{d}\beta_i \boldsymbol{c}_i,
    \label{eq:temporal_attention}
\end{equation}
It's worth noting that new geometric subspaces could be easily incorporated into Equation \ref{eq:temporal_attention} given shared real and imaginary vectors and appropriate scoring vectors.

\section{Theoretical Analysis on Temporal Patterns}
\label{sec:temporal_patterns}
Knowledge graphs exhibit \emph{patterns}. 
A \emph{structural pattern} is a regularity in the graph, e.g.\ a tree as given in the middle of Figure~\ref{fig:baseMotivExamp}, that may or may not allow for logical conclusions, but which may be hard to represent in some embedding methods. 
A \emph{logical pattern} represents a rule that allows for concluding new facts when applied to given facts. For instance, \textit{(Charles,marriedWith,Camilla)} implies
 \textit{(Camilla,marriedWith,Charles)} because \textit{marriedWith} is symmetric. 

Embeddings for temporal knowledge graphs must account for temporal facts including time components and express corresponding \emph{temporal patterns}. 
Four kinds of logical patterns, \emph{symmetric, inverse, asymmetric} and \emph{evolve} are mostly considered and studied in existing TKGE models \cite{chen2022rotateqvs,xu2020tero}. 
However, their definitions either neglect time information or merely consider patterns when facts happen at the same time. 
We generalize and go beyond these approaches and consider \emph{static temporal patterns} and \emph{dynamic temporal patterns}. If a structural or a logical temporal pattern holds \emph{regardless of time information} as in traditional knowledge graphs, we call it a \emph{static temporal pattern}. 
If a structural or a logical temporal pattern represents or draws conclusions \emph{using time information}, we call it a \emph{dynamic temporal pattern}.

In the following, we will formally define a few temporal patterns. For simplicity, we only illustrate the occasion when $\tau$ is a time interval. However, it's convenient to extend the following definitions when $\tau$ is a time point. Examples of each definition are indicated after ``//''.

\subsection{\textbf{Static Logical Temporal Patterns}} 
\begin{definition}
\label{temporalsym}
A temporal relation $p$ is symmetric at all points in time iff $\forall s,o,\tau:(s,p,o,\tau) \rightarrow (o,p,s,\tau)$. \hfill // marriedWith \\ 
A temporal relation $p$ is anti-symmetric at all points in time iff
$\forall s,o,\tau:(s,p,o,\tau) \rightarrow \neg (o,p,s,\tau). $ \hfill // locatedIn 
\end{definition}
\begin{definition}
A temporal relation $p_1$ is the inverse of temporal relation $p_2$ at all points in time 
iff $\forall s,o,\tau:$
$(s,p_1,o,\tau) \rightarrow (o,p_2,s,\tau)$. \hfill // advises, advisedBy
\label{temporalinv}
\end{definition}

\subsection{\textbf{Dynamic Logical Temporal Patterns}}
\begin{definition}
\label{dynamicsym} 
A temporal relation $p$ is temporal symmetric \\ iff $\forall s,o, \tau_1: \exists\tau_2:(s,p,o,\tau_{1}) \rightarrow (o,p,s,\tau_{2})$. \hfill // consults \\ 
A temporal relation $p$ is temporal anti-symmetric\\ iff
$\forall s,o: \exists\tau_1:(s,p,o,\tau_{1}) \rightarrow \forall \tau_2 \neg (o,p,s,\tau_{2}). $ \hfill // arrest 
\end{definition}

\begin{definition}
A relation $p_1$ at time $\tau_1$ is the temporal inverse of relation $p_2$ at time $\tau_2$ \\ iff $\forall s,o: \exists\tau_1,\tau_2: (s,p_1,o,\tau_1) \rightarrow (o,p_2,s,\tau_2)$. \hfill // invitesToVisit, Visit
\label{delayedtemporalinv}
\end{definition}


\begin{definition}
Relation $p_1$ evolves into relation  $p_2$
\\ iff $ \forall s,o: \exists  \tau_1, \tau_2: Precedes(\tau_1, \tau_2) \And (s,p_1, o, \tau_1) \rightarrow (s,p_2, o, \tau_2)$. \hfill // engagedWith, marriedWith
\label{evolvepattern}
\end{definition}


\begin{definition}\label{temprelationt1t2}
Relation $p$ is temporary in time \\ iff 
 $ \forall s,o, \tau_1: (s,p, o, \tau_1) \rightarrow \exists \tau_0, \tau_2: Precedes(\tau_0, \tau_1) \And Precedes(\tau_1, \tau_2)   \And \neg (s,p, o, \tau_0)  \And \neg (s,p, o, \tau_2)$. \hfill //worksFor 
\end{definition} 




\subsection{Modeling Temporal Patterns}
We present a theoretical analysis corresponding to the ability of our method in modeling various temporal patterns introduced in \ref{sec:temporal_patterns} as follows: (See details in Appendix F)
\begin{proposition}
HGE can model (anti-)symmetry and temporal (anti-)symmetry in Definitions \ref{temporalsym} and \ref{dynamicsym}. 
\label{th1:tempsym}
\end{proposition}

\begin{proposition}
HGE can model inverse and temporal inverse patterns in Definitions \ref{temporalinv} and~\ref{delayedtemporalinv}. 
\label{th1:tempinv}
\end{proposition}

\begin{proposition}
HGE can model evolves pattern in Definition \ref{evolvepattern}. 
\label{th1:tempevolv}
\end{proposition}

\begin{proposition}
HGE can model temporary relations in Definition \ref{temprelationt1t2}. 
\label{th1:tempevolv}
\end{proposition}




\section{Experiments}

    

\subsection{Experimental Settings}
\subsubsection{Dataset}
To evaluate the effectiveness of the proposed attention-based product space embedding, 
we perform the link prediction task on four popular temporal knowledge graph benchmark datasets, i.e.\ ICEWS14~\cite{garcia2018learning}, ICEWS05-15~\cite{garcia2018learning}, GDELT~\cite{trivedi2017know} and Wikidata12k~\cite{tcomplexlacroix2020tensor}. ICEWS14 and ICEWS05-15 are two subset datasets from the Integrated Conflict EarlyWarning System (ICEWS)\cite{lautenschlager2015icews}, which contain news facts in 2014 and between 2005 and 2015 respectively. The Global Database of Events, Language, and Tone (GDELT) is a large knowledge graph that describes facts about human behaviors. We adopt the same data subset as \cite{gao2020rotate3d}, which uses the subset of facts from April 1, 2015 to March 31, 2016. Compared to other datasets, GDELT contains fewer temporal relations but more quadruples, which makes it the densest dataset concerning temporal information. Wikidata12k is a subset of wikidata dump \cite{erxleben2014introducing}. It represents the time information $\tau \in \mathcal{T}$ as time intervals, in which $m$ or $n$ could be empty, referring to intervals $(-\infty,n]$ or $[m,\infty)$.  
Table 5 summarises the statistics of four datasets.

\subsubsection{Backbone and Baseline Models}
Our proposed model, HGE, aims to generalize complex-space-based TKGE models to an attention-based product space of heterogeneous geometric subspaces. Hence, we choose several state-of-the-art complex-space-based TKGE models as HGE's backbone models to validate its effectiveness. \textbf{TeRo} \cite{xu2020tero} defines the evolution of entity embeddings from the initial state to the current time as a rotation in complex vector space. \textbf{TComplEx} and \textbf{TNTComplEx} \cite{tcomplexlacroix2020tensor} models temporal knowledge graph completion as an order 4 tensor completion problem. \textbf{TLT-KGE} \cite{tltcomplexzhang2022along} models semantic
information and temporal information as different parts of complex space or quaternion space. 
Complex or quaternion operations exchange information between different
parts.

To give a comprehensive overview, we also compare our model with non-complex space temporal knowledge graph embedding baselines TTransE \cite{garcia2018learning},  TA-DistMult \cite{leblay2018deriving}, RotateQVS\cite{chen2022rotateqvs}, BoxTE \cite{messner2022temporal}, and LCGE\cite{niu2022logic}\footnote{We notice some inconsistent inference issues in LCGE's original code. Please refer to Appendix J for detailed discussions.}.


\subsubsection{Evaluation Metrics}
We adopt the link prediction task to evaluate our proposed model. Link prediction infers the missing entities for incomplete facts. 
During the test step, we follow the procedure of \cite{xu2020tero} to generate candidate quadruples. From a test quadruple $(s,p,o, \tau)$, we replace $s$ with $\bar{s} \in \mathcal{E}$ and $o$ with $\bar{o} \in \mathcal{E}$ to get candidate quadruples $(s,p,\bar{o}, \tau) \cup (\bar{s},p,o, \tau)$.  If $\tau$ is a time interval $[m,n]$, we sample a time point (appearing in the dataset) uniformly at random, in the range $[m,n]$ as \cite{lacroix2019tensor}. When $m$ or $n$ is empty, we set it as the first or last time point of the dataset. All candidate quadruples will be ranked by their scores using a time-aware filtering strategy \cite{goel2020diachronic}. We evaluate our models with four metrics: Mean Reciprocal Rank (MRR), the mean of the reciprocals of predicted ranks of correct quadruples, and Hits@(1/3/10), the percentage of ranks not higher than 1/3/10. For all experiments, the higher the better. 

To have a fair comparison, we set entity and relation embedding dimension sizes as reported in the original papers. For TeRo-based models, we set the dimension size of $d$ as 500 on four benchmark datasets. For TComplEx-based, TNTComplEx-based, and TLT-KGE-based models, we set the dimension size of $d$ as 1200, 1200, 1500 and 2000 on ICEWS14, ICEWS05-15, GDELT and Wikidata12k respectively. The training epoch is set to 200. We adopt the same regularizer, loss function, and negative sampling size as reported in the original papers\footnote{The code, details of training and appendix are provided in \url{https://github.com/NacyNiko/HGE}}. 

\begin{table*}[t!]
\centering
    \caption{
    Link prediction results on ICEWS14, ICEWS05-15, and GDELT. The best results among all models are in \textbf{bold}. Additionally, we \underline{underline} the best results among models with the same backbone model. 
    }
    \label{tbl:LinkPredictionResults}
\resizebox{\textwidth}{!}{ 
\begin{tabular}{lcccccccccccc}
    \hline
 \multirow{2}{*}{Model} &\multicolumn{4}{c}{ICEWS14}&\multicolumn{4}{c}{ICEWS05-15}&\multicolumn{4}{c}{GDELT}\cr 
        &MRR &Hits@1 &Hits@3 &Hits@10 &MRR &Hits@1 &Hits@3 &Hits@10 &MRR &Hits@1 &Hits@3 &Hits@10 \cr
\hline
TTransE&25.5&7.4& -&60.1&27.1&8.4& -&61.6& 11.5&0&16.0&31.8\cr
TADistMult&47.7&36.3&-
&68.6&47.4&34.6&- &72.8&20.6&12.4&21.9&36.5 \cr
RotateQVS&59.1 &50.7 & 64.2& 75.4& 63.3& 52.9& 70.9&81.3&27.0&17.5&29.3&45.8 \cr
BoxTE(k=2) & 61.5 & 53.2& 66.7& 76.7&66.4& 57.6& 72.0&82.2&33.9&25.1&36.6&50.7 \cr
LCGE & 61.6 &53.2 &66.7 &77.5 & 61.8 &51.4 & 68.1 &81.2 &-&-&-&-\cr
  \hline
 TeRo &56.2 &46.8 &62.1 &73.2 & \underline{58.6}&\underline{46.9} &\underline{66.8} &79.5&23.2&14.5&24.9&30.9 \cr
TeRo\textbf{+HGE} & \underline{58.6}& \underline{49.5}&\underline{64.5} &\underline{74.9} & 57.8&45.3 &66.5 &\underline{80.4} & \underline{23.4}&\underline{14.7}&\underline{25.2}&\underline{40.5} \cr
 $\triangle$ Improve & 4.3\%& 5.8\%& 3.9\%& 1.4\%& -1.5\% &-3.4\% & -0.1\% & 1.1\% & 0.9\% & 1.4\% &1.2\% &31.1\% \cr 
        \hline
TComplEx  &61.9 & 54.2 &66.1 &76.7 &66.5 &58.3 &71.6 &81.1&34.6&25.9&37.2&51.5 \cr
TComplEx\textbf{+HGE} &\underline{62.6} & \underline{54.7}& \underline{67.2}&\underline{77.4} & \underline{67.2}&\underline{59.3} & \underline{72.0} &\underline{81.7}& \underline{36.8} &\underline{27.4}&\underline{40.1}&\underline{55.3}  \cr

 $\triangle$ Improve & 1.1\%& 0.9\%& 1.7\%& 0.9\%& 1.1\% &1.7\% & 0.6\% & 0.7\% & 5.2\% & 5.8\% &7.8\% &7.4\% \cr 
  \hline
TNTComplEx &60.7 &51.9 &65.9 &77.2 &66.6 &58.3 &71.8 &81.7&34.1&25.2&36.8&51.5 \cr
TNTComplEx\textbf{+HGE} &\underline{63.0} &\textbf{55.1} &\underline{67.5} &\underline{78.0} &\underline{68.1} &\underline{60.1}  &\underline{72.9} &\underline{82.9} &\textbf{37.1}&\textbf{28.3}&\underline{40.0}&\underline{54.1}\cr
 $\triangle$ Improve & 3.7\%& 6.2\%& 2.4\%& 0.6\%& 2.3\% &3.1\% & 1.5\% & 1.5\% & 8.8\% & 12.3\% &8.7\% &5.0\% \cr 
        \hline
 TLT-KGE  &63.0 &54.9 &67.8 &77.7 &68.6 &60.7 &73.5 &83.1&35.6&26.7&38.5&53.2 \cr
TLT-KGE\textbf{+HGE} &\textbf{63.4} &\underline{55.0} &\textbf{68.5} &\textbf{78.8} &\textbf{68.8}&\textbf{60.8} & \textbf{74.0}  &\textbf{83.5}&\textbf{37.1}&\underline{27.7}&\textbf{40.2}&\textbf{55.6} \cr
 $\triangle$ Improve & 0.6\%& 0.1\%& 1.0\%& 1.4\%& 0.3\% &0.2\% & 1.4\% &0.5\% & 4.2\% & 3.7\% &4.4\% &3.0\% \cr 
    \hline
\end{tabular}
}
\end{table*}

\subsection{HGE's Performance Comparison}
We evaluate HGE's performance gain on four datasets. Table \ref{tbl:LinkPredictionResults} shows the performances of the original backbones and backbones plugged with HGE on time point datasets ICEWS14, ICEWS05-15, and GDELT. From Table \ref{tbl:LinkPredictionResults}, we have the following observations:

(i) HGE can provide significant improvements over chosen backbones consistently on all datasets, which verifies the effectiveness of the proposed HGE module. 

(ii) We observe the proposed method is more effective on the dense dataset GDELT. GDELT provides more instances for each relation-timestamp pair. We conjecture it benefits the temporal-geometric attention mechanism, in which fine-grained geometric attention is influenced by both relational and temporal information. Conversely, ICEWS05-15 is the sparsest dataset. As a result, HGE dose not greatly improve the performance of backbones on ICEWS05-15 and even decreases TeRo's performance.  

(iii) We find that HGE achieves greater performance gains for TNTComplEx and TComplEx than for TLT-KGE. As the TLT-KGE model provides interactions between time information and relation information in complex numbers, we believe it substitutes the function of the temporal-relation attention mechanism to some degree. However, Table 7 in Appendix  presents that TNTComplEx+HGE reaches comparable results as TLT-KGE with only half parameter numbers, which demonstrates the proposed temporal-relation attention mechanism is more efficient to combine time and relation information.

Table \ref{table:wikidata} shows link prediction results on the time interval dataset. With HGE, all metrics get improvement, reflecting HGE could boost the performance of backbones on different kinds of TKGs.

\begin{table}
\centering
    \caption{
    Link Prediction results on Wikidata12k. 
    }
    \label{table:wikidata}
    \resizebox{0.5\textwidth}{!}
{ 
\begin{tabular}{lccccc}
    \hline
  Model &MRR$[a,b]$&MRR$[a,\infty)$&MRR$(-\infty, b]$ \cr
  \hline
TNTComplEx & 27.4 & 37.8 &51.7 & \cr
TNTComplEx +HGE & 28.4 & 37.8& 57.0 & \cr
    \hline
TLT-KGE & 27.0 & 36.0& 48.0 & \cr
TLT-KGE +HGE & 27.4 & 37.7 &51.7 & \cr 
\hline
\end{tabular}
}
\end{table}

\subsection{Ablation Study}
We conduct ablation study experiments on backbone TNTComplEx to investigate the effectiveness of each component. From Table \ref{table:ablation_tga}, we have the following observations:


(i) Our proposed subspace integration strategy achieves higher performance than the stacking strategy introduced by \cite{han2020dyernie}. We find out individual loss for each subspace in TNTComplEx+stack becomes unbalanced during training time. We conjecture the model may pay too much attention to optimizing the unsuitable geometry subspaces for certain facts and hamper further improvement. 

(ii) We observe that the temporal-relation attention mechanism contributes more performance gain on GDELT. GDELT is a dense dataset and has more facts for the enumeration of objects of relation types and timestamps than other datasets. We conjecture it benefits from the fine-grained geometric attention mechanism in which the attention weights are influenced by both relation type and timestamps. 

(iii) We find that the temporal-geometric attention mechanism is more effective on ICEWS14 and ICEWS05-15 datasets. Compared to GDELT, they contain more relation types and thus provide a wider variety of relational structural patterns in the datasets. This illustrates the importance of introducing heterogeneous geometric spaces in HGE to represent the diverse structure in temporal knowledge graphs. 

\begin{table}[t!]
\centering
    \caption{
    MRR performance of HGE components. +tra stands for merely using temporal-relational attention mechanism. +tga stands for merely using temporal-geometric attention mechanism. +stack stands for integrating subspaces with the stacking strategy in \cite{han2020dyernie}
    }
    \label{table:ablation_tga}
 \resizebox{0.5\textwidth}{!}{ 
\begin{tabular}{lcccccccccccc}
    \hline
 Model &{ICEWS14}&{ICEWS05-15}&{GDELT}\cr 
 \hline
 TNTComplEx+HGE & 63.0 & 68.1& 37.1\cr
 \hline
 TNTComplEx & 60.7& 66.6 & 34.1 \cr
 TNTComplEX+stack&62.0 & 67.3 & 35.6\cr
 TNTComplEx+tra & 62.0 & 67.4 & 36.9 \cr
TNTComplEx+tga &62.6&67.5&36.4\cr
\hline
\end{tabular}
}
\end{table}

\subsection{Case Study}
\begin{figure}[t]
    \centering
    \includegraphics[width=0.45\textwidth]{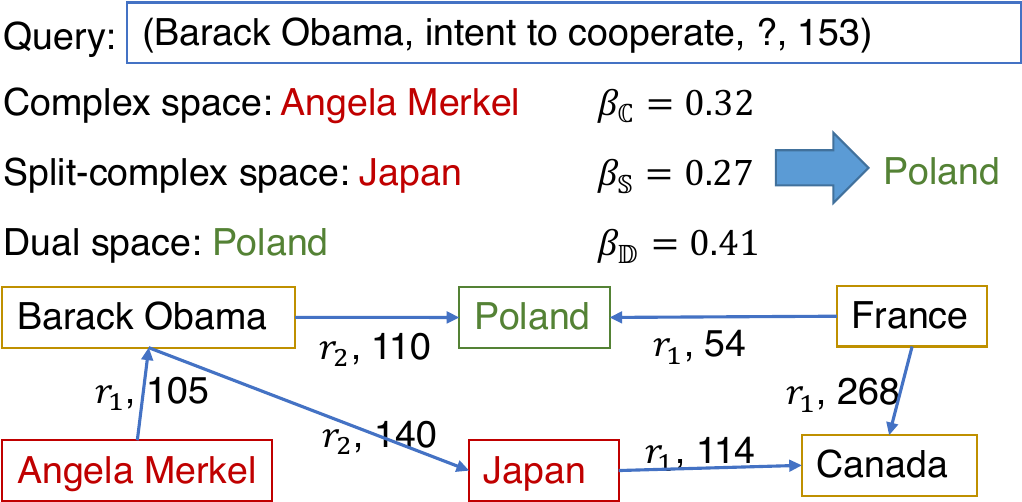}
    \caption{A case study of HGE model. We omit some entities connected to France by relation $r_{1}$ which forms a temporal star structure for brevity. $r_{1}$ stands for \textit{intent to cooperate} relation and $r_{2}$ stands for \textit{consult} relation. Time information is shown in ids. 
    }
    \label{fig:case_study}
\end{figure}

\textit{Intent to cooperate} relation forms a temporal-star structure in TKGs as the head entity could express this attitude to multiple tail entities. In Figure \ref{fig:case_study}, on account of the query (Barack Obama, intent to cooperate, ?, 153), complex space predicts the wrong answer \textit{Angela Merkel} as it supposes a symmetric instance exists for \textit{(Angela Merkel, $r_1$, Barack Obama, 105)}. Split-complex space predicts the wrong answer \textit{Japan} to form a hierarchy path between \textit{Angela Merkel, Barack Obama} and \textit{Japan}. Dual space predicts the correct answer \textit{Poland} as it has been the object entity in the temporal star structure formed by \textit{France}. Given that \textit{Barack Obama} consults \textit{Japan} recently, HGE chooses the correct answer \textit{Poland} with the help of the temporal-geometric attention mechanism.  

\section{Related Works}
TKGE models incorporate time information in different ways.
TTransE \cite{leblay2018deriving} and TA-DistMult \cite{garcia2018learning} insert the time information into different score functions as another element. 
TeRo \cite{xu2020tero} defines the temporal evolution of entity embeddings as a rotation from the initial time to the current time in complex vector space.

T(NT)ComplEx \cite{lacroix2019tensor} is a semantic matching approach that models temporal knowledge graph completion as an order 4 tensor completion problem. 
TeLM \cite{xu2021temporal} also performs 4th-order tensor factorization on temporal knowledge graphs but adds a bias component between the neighboring temporal embeddings in the temporal regularizer. Moreover, it adopts multivector embeddings for entities, relations, and timestamps.

Inspired by TeRo \cite{xu2020tero}, RotateQVS \cite{chen2022rotateqvs} embeds entities in quaternion space and temporal changes are represented as rotations.
BoxTE \cite{messner2022temporal} extends BoxE \cite{abboud2020boxe} by including  relation-specific time embeddings.
TLT-KGE \cite{tltcomplexzhang2022along} models semantic information and temporal information as different parts of complex space or quaternion space. Complex or quaternion operations exchange information between different parts. 
LCGE \cite{niu2022logic} use temporal rules to regularize entity embedding and adopts commonsense reasoning as the extra learning task. Most of the reviewed TKGE approaches model temporal patterns by using a single geometry, and do not present multiple geometries to capture diverse temporal patterns. 

Several manifold-based TKGE models have been proposed in \cite{herculemontella2021hyperbolic,han2020dyernie}.
\cite{herculemontella2021hyperbolic} is an extension of AttH \cite{chami2020low} to temporal KGEs which use hyperbolic manifolds as embedding space. It only uses a single geometry for embedding space.
\cite{han2020dyernie} embeds TKGs into a product space of several manifolds to model multiple structural patterns. However, it does not select the most suitable manifold depending on structural patterns existing in TKGs but chooses it manually.



\section{Conclusion}
We present HGE, a new temporal KGE model that utilizes multiple geometries. HGE extends state-of-the-art TKGEs from a Complex space to the product space that embeds temporal facts in Complex, Split-complex, and Dual subspaces via two temporal attention mechanisms. The temporal-relational attention mechanism captures relations with varying change frequencies. The temporal geometric attention mechanism fuses information from different geometries according to the captured relational and temporal information. Extensive experiments on benchmark datasets validate that our model uniformly improves several state-of-the-art Complex-based TKGE models. In the future, we plan to include more types of heterogeneous geometric spaces. 


\section*{Acknowledgments}
This research was funded by the German Research Foundation (DFG) via grant agreement number STA 572/18-1 (Open Argument Mining) and the German Federal Ministry for Economic Affairs and Climate Action under Grant Agreement Number 01MK20008F (Service-Meister). We would also like to thank the valuable advice from Daniel Hernández, Le Chen, Shutong Feng, and Yaxi Hu.

\bibliography{aaai24}

\begin{thebibliography}{27}
\providecommand{\natexlab}[1]{#1}

\bibitem[{Abboud et~al.(2020)Abboud, Ceylan, Lukasiewicz, and Salvatori}]{abboud2020boxe}
Abboud, R.; Ceylan, I.; Lukasiewicz, T.; and Salvatori, T. 2020.
\newblock Boxe: A box embedding model for knowledge base completion.
\newblock \emph{Advances in Neural Information Processing Systems}, 33: 9649--9661.

\bibitem[{Allen(1983)}]{allen1983maintaining}
Allen, J.~F. 1983.
\newblock Maintaining knowledge about temporal intervals.
\newblock \emph{Communications of the ACM}, 26(11): 832--843.

\bibitem[{Angeles(1998)}]{dualangeles1998application}
Angeles, J. 1998.
\newblock The application of dual algebra to kinematic analysis.
\newblock In \emph{Computational methods in mechanical systems}, 3--32. Springer.

\bibitem[{Balazevic, Allen, and Hospedales(2019)}]{balazevic2019multi}
Balazevic, I.; Allen, C.; and Hospedales, T. 2019.
\newblock Multi-relational poincar{\'e} graph embeddings.
\newblock \emph{Advances in Neural Information Processing Systems}, 32.

\bibitem[{Chami et~al.(2020)Chami, Wolf, Juan, Sala, Ravi, and R{\'e}}]{chami2020low}
Chami, I.; Wolf, A.; Juan, D.-C.; Sala, F.; Ravi, S.; and R{\'e}, C. 2020.
\newblock Low-Dimensional Hyperbolic Knowledge Graph Embeddings.
\newblock In \emph{Proceedings of the 58th Annual Meeting of the Association for Computational Linguistics}, 6901--6914.

\bibitem[{Chen et~al.(2022)Chen, Wang, Li, and Li}]{chen2022rotateqvs}
Chen, K.; Wang, Y.; Li, Y.; and Li, A. 2022.
\newblock RotateQVS: Representing Temporal Information as Rotations in Quaternion Vector Space for Temporal Knowledge Graph Completion.
\newblock In \emph{Proceedings of the 60th Annual Meeting of the Association for Computational Linguistics (Volume 1: Long Papers)}, 5843--5857.

\bibitem[{Erxleben et~al.(2014)Erxleben, G{\"u}nther, Kr{\"o}tzsch, Mendez, and Vrande{\v{c}}i{\'c}}]{erxleben2014introducing}
Erxleben, F.; G{\"u}nther, M.; Kr{\"o}tzsch, M.; Mendez, J.; and Vrande{\v{c}}i{\'c}, D. 2014.
\newblock Introducing wikidata to the linked data web.
\newblock In \emph{The Semantic Web--ISWC 2014: 13th International Semantic Web Conference, Riva del Garda, Italy, October 19-23, 2014. Proceedings, Part I 13}, 50--65. Springer.

\bibitem[{Gao et~al.(2020)Gao, Sun, Shan, Lin, and Wang}]{gao2020rotate3d}
Gao, C.; Sun, C.; Shan, L.; Lin, L.; and Wang, M. 2020.
\newblock Rotate3d: Representing relations as rotations in three-dimensional space for knowledge graph embedding.
\newblock In \emph{Proceedings of the 29th ACM International Conference on Information \& Knowledge Management}, 385--394.

\bibitem[{Garcia-Duran, Duman{\v{c}}i{\'c}, and Niepert(2018)}]{garcia2018learning}
Garcia-Duran, A.; Duman{\v{c}}i{\'c}, S.; and Niepert, M. 2018.
\newblock Learning Sequence Encoders for Temporal Knowledge Graph Completion.
\newblock In \emph{Proceedings of the 2018 Conference on Empirical Methods in Natural Language Processing}, 4816--4821.

\bibitem[{Goel et~al.(2020)Goel, Kazemi, Brubaker, and Poupart}]{goel2020diachronic}
Goel, R.; Kazemi, S.~M.; Brubaker, M.; and Poupart, P. 2020.
\newblock Diachronic embedding for temporal knowledge graph completion.
\newblock In \emph{Proceedings of the AAAI Conference on Artificial Intelligence}, volume~34, 3988--3995.

\bibitem[{Han et~al.(2020)Han, Chen, Ma, and Tresp}]{han2020dyernie}
Han, Z.; Chen, P.; Ma, Y.; and Tresp, V. 2020.
\newblock DyERNIE: Dynamic Evolution of Riemannian Manifold Embeddings for Temporal Knowledge Graph Completion.
\newblock In \emph{Proceedings of the 2020 Conference on Empirical Methods in Natural Language Processing (EMNLP)}, 7301--7316.

\bibitem[{Harkin and Harkin(2004)}]{harkin2004geometry}
Harkin, A.~A.; and Harkin, J.~B. 2004.
\newblock Geometry of generalized complex numbers.
\newblock \emph{Mathematics magazine}, 77(2): 118--129.

\bibitem[{Helzer(2000)}]{metahelzer2000special}
Helzer, G. 2000.
\newblock Special relativity with acceleration.
\newblock \emph{The American Mathematical Monthly}, 107(3): 219--237.

\bibitem[{Hogan et~al.(2021)Hogan, Blomqvist, Cochez, de~Melo, Gutierrez, Kirrane, Labra~Gayo, Navigli, Neumaier, Ngonga~Ngomo et~al.}]{hogan2021knowledge}
Hogan, A.; Blomqvist, E.; Cochez, M.; de~Melo, G.; Gutierrez, C.; Kirrane, S.; Labra~Gayo, J.~E.; Navigli, R.; Neumaier, S.; Ngonga~Ngomo, A.-C.; et~al. 2021.
\newblock Knowledge Graphs.
\newblock \emph{ACM Computing Surveys}, 54(4): 1--37.

\bibitem[{Lacroix, Obozinski, and Usunier(2019)}]{lacroix2019tensor}
Lacroix, T.; Obozinski, G.; and Usunier, N. 2019.
\newblock Tensor Decompositions for Temporal Knowledge Base Completion.
\newblock In \emph{International Conference on Learning Representations}.

\bibitem[{Lacroix, Obozinski, and Usunier(2020)}]{tcomplexlacroix2020tensor}
Lacroix, T.; Obozinski, G.; and Usunier, N. 2020.
\newblock Tensor Decompositions for temporal knowledge base completion.

\bibitem[{Lautenschlager, Shellman, and Ward(2015)}]{lautenschlager2015icews}
Lautenschlager, J.; Shellman, S.; and Ward, M. 2015.
\newblock Icews event aggregations.
\newblock \emph{Harvard Dataverse}, 3(595): 28.

\bibitem[{Leblay and Chekol(2018)}]{leblay2018deriving}
Leblay, J.; and Chekol, M.~W. 2018.
\newblock Deriving validity time in knowledge graph.
\newblock In \emph{Companion Proceedings of the The Web Conference 2018}, 1771--1776.

\bibitem[{Messner, Abboud, and Ceylan(2022)}]{messner2022temporal}
Messner, J.; Abboud, R.; and Ceylan, I.~I. 2022.
\newblock Temporal knowledge graph completion using box embeddings.
\newblock In \emph{Proceedings of the AAAI Conference on Artificial Intelligence}, volume~36, 7779--7787.

\bibitem[{Montella, Barahona, and Heinecke(2021)}]{herculemontella2021hyperbolic}
Montella, S.; Barahona, L. M.~R.; and Heinecke, J. 2021.
\newblock Hyperbolic Temporal Knowledge Graph Embeddings with Relational and Time Curvatures.
\newblock In \emph{Findings of the Association for Computational Linguistics: ACL-IJCNLP 2021}, 3296--3308.

\bibitem[{Niu and Li(2023)}]{niu2022logic}
Niu, G.; and Li, B. 2023.
\newblock Logic and Commonsense-Guided Temporal Knowledge Graph Completion.
\newblock In \emph{Proceedings of the AAAI Conference on Artificial Intelligence}, volume~37, 4569--4577.

\bibitem[{Ren et~al.(2023)Ren, Bai, Xiao, and Meng}]{ren2023hierarchical}
Ren, X.; Bai, L.; Xiao, Q.; and Meng, X. 2023.
\newblock Hierarchical Self-Attention Embedding for Temporal Knowledge Graph Completion.
\newblock In \emph{Proceedings of the ACM Web Conference 2023}, 2539--2547.

\bibitem[{Singh et~al.(2023)Singh, Kaur, Gaur, and Mausam}]{singh2023neustip}
Singh, I.; Kaur, N.; Gaur, G.; and Mausam. 2023.
\newblock NeuSTIP: A Novel Neuro-Symbolic Model for Link and Time Prediction in Temporal Knowledge Graphs.
\newblock arXiv:2305.11301.

\bibitem[{Trivedi et~al.(2017)Trivedi, Dai, Wang, and Song}]{trivedi2017know}
Trivedi, R.; Dai, H.; Wang, Y.; and Song, L. 2017.
\newblock Know-evolve: Deep temporal reasoning for dynamic knowledge graphs.
\newblock In \emph{international conference on machine learning}, 3462--3471. PMLR.

\bibitem[{Xu et~al.(2021)Xu, Chen, Nayyeri, and Lehmann}]{xu2021temporal}
Xu, C.; Chen, Y.-Y.; Nayyeri, M.; and Lehmann, J. 2021.
\newblock Temporal knowledge graph completion using a linear temporal regularizer and multivector embeddings.
\newblock In \emph{Proceedings of the 2021 Conference of the North American Chapter of the Association for Computational Linguistics: Human Language Technologies}, 2569--2578.

\bibitem[{Xu et~al.(2020)Xu, Nayyeri, Alkhoury, Yazdi, and Lehmann}]{xu2020tero}
Xu, C.; Nayyeri, M.; Alkhoury, F.; Yazdi, H.~S.; and Lehmann, J. 2020.
\newblock TeRo: A Time-aware Knowledge Graph Embedding via Temporal Rotation.
\newblock In \emph{Proceedings of the 28th International Conference on Computational Linguistics}, 1583--1593.

\bibitem[{Zhang et~al.(2022)Zhang, Zhang, Ao, Zhuang, Xu, and He}]{tltcomplexzhang2022along}
Zhang, F.; Zhang, Z.; Ao, X.; Zhuang, F.; Xu, Y.; and He, Q. 2022.
\newblock Along the Time: Timeline-traced Embedding for Temporal Knowledge Graph Completion.
\newblock In \emph{Proceedings of the 31st ACM International Conference on Information \& Knowledge Management}, 2529--2538.

\end{thebibliography}

\clearpage
\appendix

\section{Allen's Relations}
\label{Allen's Relations}
\begin{figure}[!h]
  \centering
  \includegraphics[width=0.5\textwidth]{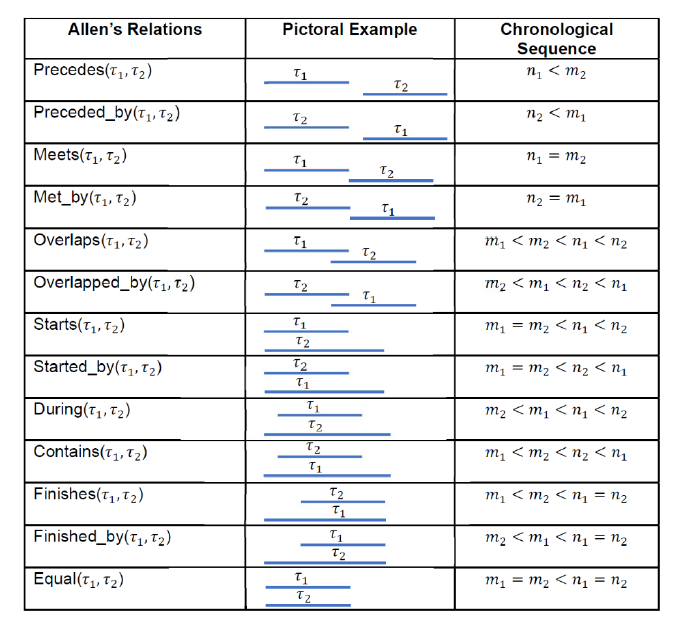}
  \caption{13 relations in Allen algebra calculus.}
  \label{fig:allen relations}
\end{figure}

The definitions of 13 Allen’s Relations between two-time intervals $\tau_1=[m_1,n_1], \tau_2=[m_2,n_2]$ are defined in Figure \ref{fig:allen relations}.

\section{Extending Backbones to HGE methods}
\paragraph{TeRo}
TeRo \cite{xu2020tero} represents $\boldsymbol{s,p,o, \tau}$ in Complex space as Equation \ref{eq:complex}. Similarly, we represent $\boldsymbol{s,p,o, \tau}$ in Split-complex space and Dual space as Equation \ref{eq:split} and \ref{eq:dual}. Following \cite{xu2020tero}, we represent $\boldsymbol{s}_{t}$ and $\boldsymbol{o}_{t}$ as:
\begin{equation}
\begin{aligned}
        \boldsymbol{s_{t}}_{\mathbb{M}_{i}}= \boldsymbol{s}_{\mathbb{M}_{i}} \circ \boldsymbol{t}_{\mathbb{M}_{i}}, 
                \boldsymbol{o_{t}}_{\mathbb{M}_{i}}= \boldsymbol{o}_{\mathbb{M}_{i}} \circ \boldsymbol{t}_{\mathbb{M}_{i}}
\end{aligned}
\end{equation}
For temporal relational attention, we set $\boldsymbol{p} = \boldsymbol{p}_{s} = \boldsymbol{p}_{c}$, so the dynamic relational information is captured by $\boldsymbol{p}*\boldsymbol{t}$.

\paragraph{TLT-KGE}
TLT-KGE \cite{tltcomplexzhang2022along} represents $\boldsymbol{s,p,o, \tau}$ in Complex space as:
\begin{equation}
\begin{aligned}
        \boldsymbol{s_{\mathbb{C}}}=\boldsymbol{e}_{s} + \boldsymbol{t}_{\tau}^{e}i, \boldsymbol{p_{\mathbb{C}}}=\boldsymbol{r}_{p} + \boldsymbol{t}_{\tau}^{r}i,  \boldsymbol{o_{\mathbb{C}}}=\boldsymbol{e}_{o} + \boldsymbol{t}_{\tau}^{e}i, 
\end{aligned}
\end{equation}
Similarly, we represent $\boldsymbol{s,p,o, \tau}$ in Dual space and Split-complex space as:
\begin{equation}
\begin{aligned}
        \boldsymbol{s_{\mathbb{D}}}=\boldsymbol{e}_{s} + \boldsymbol{t}_{\tau}^{e}\epsilon, \boldsymbol{p_{\mathbb{C}}}=\boldsymbol{r}_{p} + \boldsymbol{t}_{\tau}^{r}\epsilon,  \boldsymbol{o_{\mathbb{C}}}=\boldsymbol{e}_{o} + \boldsymbol{t}_{\tau}^{e}\epsilon, \\
        \boldsymbol{s_{\mathbb{S}}}=\boldsymbol{e}_{s} + \boldsymbol{t}_{\tau}^{e}j, \boldsymbol{p_{\mathbb{S}}}=\boldsymbol{r}_{p} + \boldsymbol{t}_{\tau}^{r}j,  \boldsymbol{o_{\mathbb{S}}}=\boldsymbol{e}_{o} + \boldsymbol{t}_{\tau}^{e}j, 
\end{aligned}
\end{equation}
For temporal relational attention, we adopt $\boldsymbol{r_{p}}$ and $\boldsymbol{r_{compr}}$ in Equation 12 of original paper as  $\boldsymbol{p}_{s}$, and $\boldsymbol{p}_{c}*\boldsymbol{\tau}_{\tau}$ respectively, where $\boldsymbol{r_{compr}}$ = $\boldsymbol{r_{p}}$ *  $\boldsymbol{t_{compr}}$.  

\section{Embeddings in Complex, Dual and Split-Complex Subpaces}
\label{sec:embeddings_subspaces}
Keeping other settings fixed, we train 3 model variants TNTComplEx+complex, TNTComplEx+split, TNTComplEx+dual which use a single geometric space to optimal MRR scores on ICEWS14. We randomly select 100 entities from the entity set and analyze the similarity of their embeddings on different geometric spaces by cosine similarity:

\begin{equation}
S_C(A, B)=\frac{\mathbf{A} \cdot \mathbf{B}}{\|\mathbf{A}\|\|\mathbf{B}\|}=\frac{\sum_{i=1}^n A_i B_i}{\sqrt{\sum_{i=1}^n A_i^2 \cdot \sum_{i=1}^n B_i^2}}
\end{equation}

\begin{figure}
     \centering
     \begin{subfigure}[b]{0.48\textwidth}
         \centering
          \includegraphics[width=\textwidth]{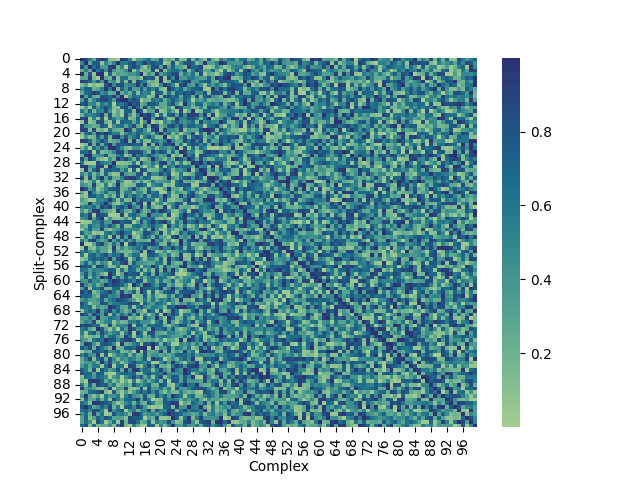}
         \caption{Cosine similarity score between trained entity embeddings in Complex space and Split-complex space.}
         \label{fig:complex_split}
     \end{subfigure}
     \hfill
     \begin{subfigure}[b]{0.48\textwidth}
         \centering
        \includegraphics[width=\textwidth]{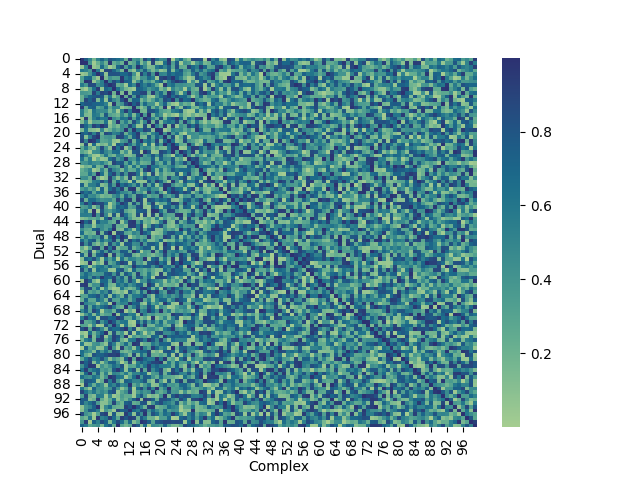}
         \caption{Cosine similarity score between trained entity embeddings in Complex space and Split-complex space.}
         \label{fig:complex_dual}
     \end{subfigure}
     \hfill
          \begin{subfigure}[b]{0.48\textwidth}
         \centering
          \includegraphics[width=\textwidth]{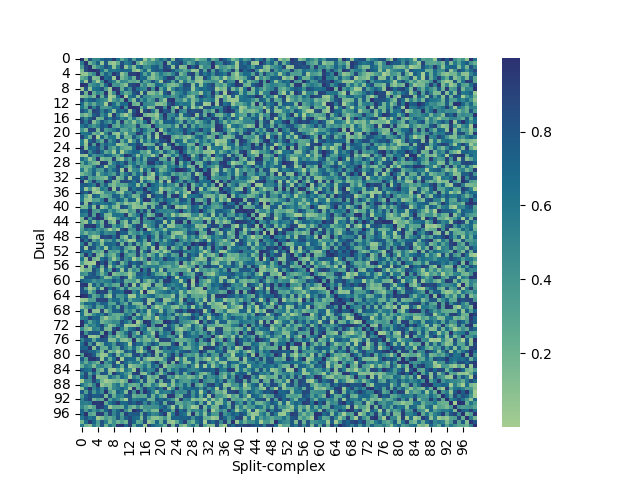}
         \caption{Cosine similarity score between trained entity embeddings in Split-complex space and Dual space.}
         \label{fig:split_dual}
     \end{subfigure}
     \hfill
        \caption{Cosine similarity scores between entity embeddings from different geometric space. x-axis and y-axis show the entity id on relevant geometric space}
        \label{fig:cosine_similarity}
\end{figure}

We concat the real part and imaginary part of one entity when calculating the cosine similarity. From Figure \ref{fig:cosine_similarity}, we could find out that in every sub-graph, the values on the diagonal, which represent the cosine similarity between entity embeddings of the same entity on different geometric subspace, are always the highest in a row and exceed 0.95. Therefore, although their geometric interpretations are diverse, real and imaginary vectors in different subspaces are almost unanimous when trained to optimal settings.

\section{Dataset Overview}
\label{sec:dataset}

Dataset statistics are described in Table \ref{table:dataset}. 

\section{Temporal Pattern Statistics}
We calculate the occurrence of each temporal pattern introduced in Section \ref{sec:temporal_patterns} to give an overview distribution of the temporal patterns. Table \ref{table:pattern} shows the statistics on ICEWS14, ICEWS05-15 and GDELT. If a group of quadruples, such as the examples shown in the 2nd column in Table \ref{table:pattern}, meets the definition in Section \ref{sec:temporal_patterns}, we calculate it as one occurrence. 

\begin{table*}
\centering
    \caption{
    Real examples and statistics of each pattern in the train set of ICEWS14, ICEWS05-15, GDELT.
    }
    \label{table:pattern} 
\resizebox{\textwidth}{!}
{\begin{tabular}{lcccc}
    \hline
 Patterns &Examples&ICEWS14&ICEWS05-15&GDELT \cr
  \hline
    \multirow{2}{*}{static symmetric}   &(Iraq, sign formal agreement, Iran, 2014-04-06)&6,506  &36,537&366,830 \cr
 &(Iran, sign formal agreement, Iraq, 2014-04-06)&&\cr
 \hline
    \multirow{2}{*}{static inverse}  &(Fiji, host a visit, Julie Bishop, 2014-11-04), &10,361  &63,092&552,280 \cr
 &(Julie Bishop, make a visit, Fiji, 2014-11-04)&&\cr
 \hline
 \multirow{2}{*}{dynamic symmetric}&(France, engage in negotiation, Poland, 2014-04-04)&78,473  &3,817,343& 17,265,293 \cr
 &(Poland, engage in negotiation, France, 2014-02-20)&&\cr
 \hline
 \multirow{2}{*}{dynamic inverse}  &(Angela Merkel, discuss by telephone, Ukraine, 2014-03-14), &768,586  &48,641,730&104,909,248 \cr
 &(Ukraine, consult, Angela Merkel, 2014-03-27)&&\cr 
 \hline
    \multirow{2}{*}{dynamic evolve}  &(South Korea, demand, Japan, 2014-07-15), &971,055  &63,733,447&112,653,245 \cr
 &(South Korea, reject judicial cooperation, Japan, 2014-07-18)&&\cr
 \hline
\end{tabular}
}
\end{table*}

\section{Modeling Various Temporal Patterns}
\label{temporal_pattern_proof}

 \begin{proposition}
HGE can model (anti-)symmetry patterns introduced in Definitions \ref{temporalsym} and \ref{dynamicsym}.
\label{th1:tempsym}
\end{proposition}
\begin{proof}
Let $p$ be a relation with temporal symmetry.
One condition for modeling this pattern is $S(s,p,\bar{o},\tau) =  S(o,p,\bar{s},\tau).$  For simplicity of representation, we use $p_t = p_{s\tau}.$
Without loss of generality, we assume that we have only a one-dimensional split-complex vector. Therefore, we have the following equality to fulfill temporal symmetry:
\begin{equation*}
\begin{split}
    &\big( s_a p_{ta} o_{a} + s_b p_{tb} o_{a} - s_a p_{tb} o_{b} - s_b p_{ta} o_{b}\big) = \\
    &\big( o_a p_{ta} s_{a} + o_b p_{tb} s_{a} - o_a p_{tb} s_{b} - o_b p_{ta} s_{b} \big).
\end{split}
\end{equation*}
This leads to the following equality $s_b p_{tb} o_{a} = s_a p_{tb} o_{b}.$
To hold this equality, we need to have either $p_{tb} = 0$ or $s_b o_{a} = s_a o_{b}$. 
So far, we show for a given grounded quadruple $(s,p,o,\tau)$, 
if our model learns $(s,p,o,\tau)$ to be true, it can also hold its temporal symmetry $(o,p,s,\tau)$ as true. 
To generalize this to the universal quantifier (every grounded quadruple),
we can add one extra dimension to model temporal symmetry for the extra pair of entities. In the extended dimension for the new pair (s,o), we should have $p_{tb} = 0$ or $s_b o_{a} = s_a o_{b}$ to hold temporal symmetry. In this way, all pairs $(s,o)$ which are connected by temporal symmetry relation will be held as true by the model.
A similar procedure can be done for Dual and ComplEx spaces. Therefore, there exist assignments for embeddings of entities and relations that fulfill the encoding of the temporal symmetric pattern. 
\end{proof}

\begin{proposition}
HGE can model inverse patterns introduced in Definitions \ref{temporalinv} and~\ref{delayedtemporalinv}.
\label{th1:tempinv}
\end{proposition}
\begin{proof}
Let temporal relation $p_1$ be the inverse of the temporal relation $p_2$ at all time points (\ref{temporalinv}).
One condition to model this pattern is to fulfill $S(s,p_1,\bar{o},\tau) =  S(o,p_2,\bar{s},\tau).$
Without loss of generality, we assume that we have only a one-dimensional split-complex vector. 
Therefore, we have the following equality to fulfill temporal inverse relationships:
\begin{equation*}
\begin{split}
    &\big( s_a p_{1ta} o_{a} + s_b p_{1tb} o_{a} - s_a p_{1tb} o_{b} - s_b p_{1ta} o_{b}\big) = \\
    &\big( o_a p_{2ta} s_{a} + o_b p_{2tb} s_{a} - o_a p_{2tb} s_{b} - o_b p_{2ta} s_{b} \big).
\end{split}
\end{equation*}
If we set $p_{1ta}  = p_{2ta},\,\,\, p_{1tb}  = -p_{2tb}$, the above equality holds. This means there exist assignments for embeddings of entities, relations, and times that fulfill the encoding of temporal inverse patterns. 
Our proof can be generalized to $d$ dimensional product space by adding one dimension per each grounded atom. For the pattern in \ref{delayedtemporalinv}, the proof procedes likewise. 
The only difference is that the time embedding will be different at the two times $\tau_1, \tau_2$ to hold $p_{1ta}  = p_{2ta},\,\,\, p_{1tb}  = -p_{2tb}$. 
\end{proof}

\begin{proposition}
Let us assume that relation $p_1$ evolve to relation $p_2$ as formalized in \ref{evolvepattern}. HGE can model this pattern.
\label{th1:tempevolv}
\end{proposition}
\begin{proof}
Given that $p_1$ evolves to $p_2$, and also given 
the two times $\tau_1$ and $\tau_2$ with $\tau_1 \prec\leq \tau_2$,
to model the pattern, we need to have $S(s,p_1,\bar{o},\tau_1) =  S(s,p_2,\bar{o},\tau_2).$
Without loss of generality, we assume that we have only a one-dimensional split-complex vector. Then, we must fulfill the following equality:
\begin{align*}
    \big( s_a p_{1t_1a} o_{a} + s_b p_{1t_1b} o_{a} - s_a p_{1t_1b} o_{b} - s_b p_{1t_1a} o_{b}\big) =  \\
    \big( s_a p_{2t_2a} o_{a} + s_b p_{2t_2b} o_{a} - s_a p_{2t_2b} o_{b} - s_b p_{2t_2a} o_{b}\big).
\end{align*}
For this equality to hold, it must be the case that  $p_{1t_1a} = p_{2t_2a},\,\,\, p_{1t_1b} = p_{2t_2b}$. Note that these equality conditions do not necessarily mean that the embedding of static and temporal relations in Equation \ref{eq:temporalAttention} should be the same because different convex combinations can create the same vector for temporal relations. Considering the  universal quantifier, we can add one extra dimension for each grounded atom to fulfill equality. 
A similar consideration can be applied to Dual and ComplEx spaces. Therefore, there exist assignments for embeddings of entities and relations that encode the patterns.
\end{proof}

\begin{proposition}
Let $p$ be a temporary relation in time as defined in \ref{temprelationt1t2}. HGE can model this relation.
\label{th1:tempevolv}
\end{proposition}

\begin{proof}
Let $p$ be a temporary relation as in \ref{temprelationt1t2}.
To follow this pattern in the embedding space, for a given grounded atom $(s,p,o,\tau_1)$, there exist $\tau_0, \tau_2$ and also the embedding vectors for $s, p, o, \tau_0, \tau_1, \tau_2$ such that we have $S(s,p,\bar{o},\tau_1) \neq  S(s,p,\bar{o},\tau_2)$ and $S(s,p,\bar{o},\tau_1) \neq  S(s,p,\bar{o},\tau_0)$ as one possible condition to fulfill the pattern.
Similar to the previous proofs, 
let us assume that we have only 
a one-dimensional split-complex vector. 
To fulfill the first condition (the second one will be similar), we have
\begin{align*}
    \big( s_a p_{t_1a} o_{a} + s_b p_{t_1b} o_{a} - s_a p_{t_1b} o_{b} - s_b p_{t_1a} o_{b}\big) \neq  \\
    \big( s_a p_{t_2a} o_{a} + s_b p_{t_2b} o_{a} - s_a p_{t_2b} o_{b} - s_b p_{t_2a} o_{b}\big).
\end{align*}
This can be simply fulfilled if we set $p_{t_1a} \neq p_{t_2a}, p_{t_1b} \neq p_{t_2b}$.
In addition, we can have a large value for $S(s,p,\bar{o},\tau_1)$ and a small value for $S(s,p,\bar{o},\tau_2)$ (or vice versa) by properly setting the temporal relation close to zero at time $\tau_1$ and high value at time $\tau_2$ (and vice versa).
A similar calculation can be done for Dual and ComplEx spaces. 
Therefore, there exist assignments for embeddings of entities and relations that  encode the pattern.
\end{proof}

\section{Experiment Details}
All experiments in the paper were run on the same NVIDIA A100 GPU device(40G GPU/100G CPU) with Ubuntu system 22.0. We implement a grid search to select the best regularizer weight from [5e-4, 3e-3, 5e-3, 3e-3, 1e-3, 3e-2, 1e-2, 1e-1]. A detailed list of hyperparamters is provided in hyperparamter.pdf file in the code folder of supplement material.

\begin{table}
\centering
    \caption{
    Statistics for ICEWS14, ICEWS05-15, GDELT and Wikidata12k.
    }
    \label{table:dataset}
    \resizebox{0.5\textwidth}{!}
{ 
\begin{tabular}{lcccc}
    \hline
 Dataset &ICEWS14&ICEWS05-15&GDELT&Wikidata12k \cr
  \hline
 Entities  &7,128 &10,488  &500 &12,554 \cr
Relations & 230 & 251  & 20& 24 \cr
Times & 365 & 4017 &366 &1,726\cr
Train& 72,826 &386,962 & 2,735,685&32,497\cr
Validation & 8,941& 46,275 & 341,961&4,062
\cr 
Test & 8,963 & 46,092 & 341,961&4,062\cr 
    \hline
\end{tabular}
}
\end{table}

\section{Temporal Structural Patterns on Geometric subspaces}
We consider symmetric patterns belonging to structural patterns too and define two other types of temporal structural patterns:
\begin{definition}
Relation $p$ forms a temporal star of size $n\in\mathbb{N}$\\
iff $\forall s: \exists o_1,\tau_1 \ldots, 
 o_n, \tau_n: Precedes(\tau_1, \tau_2) \And \ldots \And Precedes(\tau_(n-1), \tau_n) \And (s,p, o_1, \tau_1) \And (s,p, o_2, \tau_2) \And \ldots \And (s,p, o_n, \tau_n)$. 
\end{definition}

\label{sec:structural_pattern_analysis}
\begin{definition}\label{temphierarchy}
A relation $p$ forms a temporal hierarchy iff $\forall v_1, v_2, v_3, \tau_1, \tau_2: (v_1,p,v_2,\tau_1)\And (v_2,p,v_3,\tau_2) \rightarrow \tau_1 \prec \tau_2 
$ 
\end{definition}

We investigate if heterogeneous geometric subspaces could represent different kinds of structural patterns. We extract 3 subsets for static symmetry, temporal hierarchy, and temporal star structural patterns from the test set of ICEWS14 and ICEWS05-15. Four variants of TNTComplEx+HGE model are tested in these subsets: 1) complex: only complex space is used. 2) split-complex: only split-complex space is used. 3) dual: only dual space is used. 4) HGE: the full model with three heterogeneous subspaces. 


 Table \ref{table:structural patterns} shows that models using complex space perform best on static symmetric structural patterns. Models using split-complex space performs best on temporal hierarchy pattern while models using dual space perform best on temporal star pattern. This observation supports our core assumption 
that multiple geometric spaces may benefit temporal knowledge graph representation. Moreover, TNTComplEx+HGE performs better than all variants with single geometric spaces, demonstrating that the proposed product space with temporal geometric attention mechanism could integrate the advantages of individual subspaces. 

\begin{table*}[t!]
\centering
    \caption{
    MRR performance of heterogeneous geometric spaces on diverse structural pattern subsets.
    }
    \label{table:structural patterns}
{ 
\begin{tabular}{lccccccc}
    \hline
 Datasets & Structural Patterns & Statistics& TNTComplEx & complex& split-complex& dual& HGE\cr 
 \hline
 \multirow{3}{*}{ICEWS14}
 &static symmetric & 1352&98.8& \textbf{99.5} & 99.3 & 98.3 &\textbf{99.5}\cr
  & temporal hierarchy& 1193& 69.5& 70.4& \textbf{71.8} & 71.0 &\textbf{71.8}  \cr
  &temporal star &6197 & 70.5&71.6 & 71.9 & \underline{72.9}& \textbf{73.0}\cr
\hline
  \multirow{3}{*}{ICEWS05-15}
 &static symmetric & 7240& 99.7&\textbf{99.8} & 99.7 & 99.6 &\textbf{99.8}\cr
  & temporal hierarchy&16703& 72.7& 72.8& \underline{73.7}& 72.5&\textbf{74.3} \cr
  &temporal star &39724 &73.8& 73.8 & 72.4 & \underline{74.7}& \textbf{75.4}\cr
\hline
\end{tabular}
}
\end{table*}

\section{HGE's Time and Space Usage}
 As HGE reuses vectors for different geometric subspaces, the increased parameters to implement an HGE module will be $2|\mathcal{R}|*d$, which is the attention weights for two proposed attention mechanisms. We demonstrate the HGE's efficiency by comparing the number of parameters and running times of the original backbone with HGE-extended backbones. All models are trained with 200 epochs and we calculate the average running time of training epochs for each model. From Table \ref{table:time and space}, we observe that with the same embedding dimension $d$=1200 for entities and relations, the increased number of parameters and running time are rather moderate for HGE-extensions. Specifically, when TNTComplEx is extended by HGE, its performance is comparable to TLT-KGE with only half as many parameters and a shorter running time. Even if we decrease $d$ of TNTComplEx+HGE to 1100, it still outperforms backbone TNTComplEx($d$=1200) with fewer parameter numbers. This demonstrates that HGE's improvements do not come from the increased number of parameters, but rather from its representational approach.

\begin{table*}[h]
\centering
    \caption{
    Parameter number and average runtime for original backbones and backbones extended by HGE.
    }
    \label{table:time and space}
{ 
\begin{tabular}{lccccc}
    \hline
 Datasets & Model & Rank($d$) & Parameter number\ \ & \ \ Average epoch time(s) & MRR\cr 
 \hline
 \multirow{4}{*}{ICEWS14}
 & TNTComplEx&1200& 20,191,200 &1.80 & 60.7  \cr
 &TLT-KGE&1200 &38,693,400&2.25 & 63.0\cr
  \cline{2-6}
   &TNTComplEx+HGE&1100 & 19,520,600&2.10 & 62.9\cr
 &TNTComplEx+HGE&1200 & 21,295,200&2.19 & 63.0\cr
\hline
 \multirow{4}{*}{ICEWS05-15}  
 & TNTComplEx&1200& 37,221,600 & 11.79 & 66.6 \cr
 &TLT-KGE&1200 &81,360,600&13.91 &68.6 \cr
   \cline{2-6}
    &TNTComplEx+HGE&1100 & 35,224,200& 11.52& 67.7\cr
 &TNTComplEx+HGE&1200 & 38,426,400& 12.13& 68.1\cr

\hline
\end{tabular}
}
\end{table*}

\begin{table*}[!h]
\centering
    \caption{
    Results of LCGE in original paper and by our implementation
    }
    \label{table:lcge}
{\begin{tabular}{lcccccccc}
    \hline
\multirow{2}{*}{Model} &\multicolumn{4}{c}{ICEWS14}&\multicolumn{4}{c}{ICEWS05-15}\cr 
           &MRR &Hits@1 &Hits@3 &Hits@10 &MRR &Hits@1 &Hits@3 &Hits@10  \cr
\hline
LCGE\cite{niu2022logic} & 92.5&91.6& 92.9&93.7 & 91.2 &90.3 &91.6&92.5 \cr
LCGE & 61.6 &53.2 &66.7 &77.5 & 61.8 &51.4 & 68.1 &81.2\cr
    \hline
\end{tabular}
}
\end{table*}

\section{Baseline Selection}
\label{sec:baseline selection}
\paragraph{LCGE} We found out the commonsense reasoning score introduced in equation 11 of LCGE\cite{niu2022logic} was considered during the training time but missed during the test time, which causes bias to final scores and rankings. We re-implemented the codes and attached our implementation in the supplementary material's code/LCGE\_new folder. Table \ref{table:lcge} shows the comparison of reported results in \cite{niu2022logic} and results by our implementation. 

\paragraph{DyERNIE} We do not include the baseline of DyERNIE\cite{han2020dyernie} since this paper reports the results using the static filtered setting. Moreover, the code released by the authors is not complete to implement hyperbolic spaces, making it hard to report time-aware filtering results.

\paragraph{HSAE} HSAE\cite{ren2023hierarchical} adopts a hierarchy self-attention mechanism to incorporate information from different time shots. We do not include the baseline of HSAE because the author does not publish the codes.

\clearpage

\end{document}